\documentclass[journal]{IEEEtran}

\usepackage[cmex10]{amsmath}
\usepackage{epsfig}
\usepackage{amsfonts}
\usepackage{url}
\usepackage{cite}
\usepackage{multirow}
\usepackage{subfig}
\usepackage{amssymb}
\usepackage{graphics} 
\usepackage[usenames, dvipsnames]{color}
\usepackage[perpage]{footmisc}
\usepackage[sort, numbers]{natbib}
\usepackage{ifthen}
\usepackage{multirow}
\usepackage{multicol} 
\usepackage{balance}

\newcommand{\bm}[1]{\mbox{\boldmath$#1$}}
\newcommand{\vect}[1]{\bm{#1}}

\begin{document}

\title{Image interpolation using Shearlet based \\iterative refinement}

\author{H.~Lakshman, W.-Q Lim, H.~Schwarz, D.~Marpe,
G.~Kutyniok, and~T.~Wiegand,~\IEEEmembership{Fellow,~IEEE}%
\thanks{H.~Lakshman, H.~Schwarz, and D.~Marpe, are with the Image \& Video Coding group, 
Fraunhofer Institute for Telecommunications -- 
Heinrich Hertz Institute (Fraunhofer HHI), 
10587 Berlin, Germany (email:~firstname.lastname@hhi.fraunhofer.de).
W.-Q~Lim and G.~Kutyniok are affiliated with the Department of Mathematics, 
Technical University of Berlin, 10587 Berlin, Germany.
T. Wiegand is jointly affiliated with the Image Processing Department,
Fraunhofer HHI, and the Image Communication Chair, Technical University of
Berlin, 10587 Berlin, Germany.}
\thanks{Copyright (c) 2012 IEEE. Personal use of this material is permitted. 
        However, permission to use this material for any other purposes must 
        be obtained from the IEEE by sending an email to 
        pubs-permissions@ieee.org.}%
%\thanks{Manuscript received Month date, Year; revised Month date, Year.}
}

\maketitle

\begin{abstract}
This paper proposes an image interpolation algorithm exploiting sparse representation for natural images. 
It involves three main steps: 
(a)~obtaining an initial estimate of the high resolution image using linear methods like FIR filtering, 
(b)~promoting sparsity in a selected dictionary through iterative thresholding, and 
(c)~extracting high frequency information from the approximation to refine the initial estimate.
For the sparse modeling, a shearlet dictionary is chosen to yield a multiscale directional representation.
The proposed algorithm is compared to several state-of-the-art methods to assess its objective 
as well as subjective performance.
Compared to the cubic spline interpolation method, 
an average PSNR gain of around 0.8~dB is observed over a dataset of 200 images.
\end{abstract}


\IEEEpeerreviewmaketitle

\IEEEPARstart{I}{mage} interpolation refers to generating a high resolution (HR) 
image from an input low resolution (LR) image. 
The resolution of an image can be defined in various ways, e.g.,
based on:
\begin{itemize}
\item the number of pixels in the image,
\item the characteristics of the physical sensing device in the camera,
\item the effective sharpness as perceived by a human observer.
\end{itemize}

\noindent To quantify the resolution based on the first method is simple,
but the latter two are considerably more complex.

Interpolation tasks have regained attention because images/videos are 
being viewed on displays of different sizes, like mobile phones, 
tablets, laptops, PCs, etc. 
For example, the content for a 1080p display may be available 
in a 720p format and needs to be interpolated. 
More recently, 4K displays are becoming popular and 
content with a lower resolution
may have to be displayed on them. 
It also finds many applications in computer vision, 
graphics, compression, editing, surveillance
and texture mapping. 
It is vital for image browsing and video playback software. 
Details synthesis in image interpolation can also be used 
as a tool in spatial scalable video coding.

Image interpolation, due to its interdisciplinary applications, 
is referred to using various terms, including image upsampling, 
upscaling, resizing, resampling, etc.,
depending on the community one comes from. 
%The term super-resolution is used loosely in cases where some 
%\emph{extra} information is produced than what is actually 
%present in the current LR image, 
%e.g., using other LR images in the video, 
%model priors for HR images, etc. 
Many established methods are available for achieving interpolation, 
e.g., FIR filtering, spline based schemes, etc. 
These techniques are sufficient for many practical purposes,
but may cause several artifacts, most commonly, 
blurring of the resulting HR image. 
The main goal of this paper is to recover sharp edges and textures, 
while reducing blurring, ringing, aliasing or other visual artifacts
in the resulting HR images. 
For videos, there is an additional requirement to maintain the temporal 
coherence to avoid picture-to-picture flickering during playback. 

Efficient image representation is at the heart of image interpolation.
Natural images occupy only a small fraction of the entire space
of all possible images.
Images show geometric structures, like edges, and conventional
Fourier or DCT domains are not well suited for accurate modeling
or extraction of geometric structures,
although they are very useful in compression applications.

\section{State-of-the-art}
\label{sec:soa}

%Interpolation is a fundamental task in image processing
%and numerous methods exist for realizing it.
To review some important mathematical principles,
a categorization of various methods is provided here.

\subsubsection*{Linear methods}
Signal processing theory for band limited signals, 
advocates sampling higher than Nyquist rate and 
a sinc interpolation~\cite{Shannon1949, Unser2000}. 
The assumption of band limitedness does not hold for most images 
due to the existence of sharp edges.
However, conventional schemes adhere to this philosophy and 
approximate the ideal low pass filter to produce acceptable 
results for many practical applications. 
Techniques like bilinear or bicubic interpolation
are some popular examples that have very low computational complexity.
Extending the sampling theory to shift-invariant spaces without band limiting constraints
has led to a generalized interpolation framework, 
e.g., B-spline~\cite{Unser1999} and MOMS interpolation~\cite{Blu2001} 
that provide improvements in image quality for a 
given support of basis functions. 
However, these linear models cannot capture the 
fast evolving statistics around edges. 
Increasing the degree of the basis functions in these linear models 
helps to capture higher order statistics but result in longer effective support 
in the spatial domain and hence produce artifacts like ringing around edges.

\subsubsection*{Directional methods}
To improve the linear models, 
directional interpolation schemes have been proposed 
that perform interpolation along the edge directions 
instead of going across the edges.
Some schemes in this class use edge detectors~\cite{Allebach1996, Shi2002}.
The method in \emph{New edge directed interpolation} (NEDI)~\cite{Li2001} 
computes local covariances in the input image and 
uses them to adapt the interpolation at the higher resolution, 
so that the support of the interpolator is along the edges.
However, the resulting images still show some artifacts. 
%The local covariance based method is improved in~\cite{Zhang2008}
%and provides better results.
The iterative back projection~\cite{Irani1991} technique 
improves image interpolation
when the downsampling process is known.
Its basic idea is that the reconstructed HR image 
from the LR image should produce the same observed 
LR image when it is passed through the same blurring and 
downsampling process. 
However, the downsampling filter may not be known in many cases
or the input image may be camera captured, 
where the optical anti-alias filter used within the sampling system 
is not known during the subsequent image processing stages. 
Therefore, it is desirable to design a method that 
does not rely directly on the downsampling process. 

\subsubsection*{Sparsity based methods}
Image interpolation can be seen as an estimation problem
where the input data are inadequate.
Naturally, the solution to this problem is not unique due to the lack of 
information in the HR grid.
A popular idea used in such underdetermined problems is to exploit the structure of the desired
solution. %, if some statistics of the desired solution are known a priori.
For images, sparsity in transform domains has proven itself to be 
a very useful prior~\cite{Field94, Olshausen1996, Olshausen1997}.
Sparse approximation can be viewed as approximating a signal 
with only a few expansion coefficients~\cite{SelesnickSP}.
Sparsity priors have also been proposed for image interpolation,
e.g., in~\cite{Mueller2007, Yang2010, Mallat2010}.
The method in~\cite{Mueller2007} uses a contourlet transform for 
sparse approximation and 
is designed for an observation model that assumes that the LR image
is the low pass subband of a wavelet transform.
It uses the same transform in a recovery framework,
so it relies directly on knowledge of the downsampling process.
We follow a similar recovery principle, 
but design a system so that it works for typical anti-aliased LR images
instead of requiring a specific wavelet transform.
%Another set of methods use learning based approach.
The method in~\cite{Yang2010} involves jointly training two 
dictionaries for the low- and high-resolution image patches. 
The set of all elements that can be used in the expansion
is called a dictionary.
It then performs a sparsity based recovery,
but involves high search complexity to determine a sparse approximation
in the trained dictionary (observed to be more than 100x slower than~\cite{Mueller2007}).
The approach in~\cite{Mallat2010} considers the case when the LR
image produced by sub-sampling a HR image is aliased.
The method in~\cite{Dong11} learns a series of compact sub-dictionaries and 
assigns adaptively a sub-dictionary to each local patch as the sparse domain.
The K-SVD algorithm proposed in~\cite{Aharon2006} and its extensions are commonly used
for learning an overcomplete dictionary. 
These methods depend on the similarity of training and test patches
and number of the selected examples, 
which are typical issues in learning-based algorithms.
%The main advantage of using fixed transforms, 
%instead of training based, 
%is that the synthesis of HR information from LR input data 
%need not impose constraints learned from training examples. 
Furthermore, analytically determined transforms have structures
that can be exploited to produce a fast implementation,
which might be hard to impose during dictionary learning.

\subsubsection*{Discussion of the proposed method}
We recognize the fact that linear models such as 
interpolation based on FIR filters are faithful in interpolating the low frequency 
components but distort the high frequency components in the 
upsampled image. 
An iterative framework, based on~\cite{Guleryuz06a, Mueller2007},
is proposed that combines the output from an 
initial interpolator and detail components from a denoised approximation. 
The method used here for denoising is the so-called 
shrinkage or thresholding approach, i.e.,
by transforming the signal to a specific domain, 
setting the transform coefficients below a certain (absolute) value to zero and 
inverse transforming the coefficients to get back an approximation. 
The domain used for transforming is chosen so that the coefficients
with large absolute values capture most of the geometric features 
and the coefficients with low absolute values constitute noise or finer details.
To this end, multi-resolution transforms or 
multi-resolution directional transforms are preferred.
The concepts of multi-resolution and directionality in transforms
are reviewed in Sec.~\ref{sec:sr},
based on which a framework for details synthesis in interpolation 
is proposed in Sec.~\ref{sec:fw}.
%Instead of explicitly detecting the directions to interpolate, 
%thresholding of directional transform coefficients 
%produces sharper image along the dominant edges.
In fact, wavelet domain thresholding has 
been successfully applied to many denoising 
problems~\cite{Donoho94, Donoho1995}.
Due to the subsampling in orthogonal wavelet transforms,
they are not translation invariant.
But, unlike a typical compression scenario,
the number of transform coefficients generated during
modeling or denoising need not be the same as the number of input samples.
This is exploited by removing the sub-sampling in the wavelet transform
and is shown to yield better denoising 
results~\cite{Coifman1995, Fowler2005}.
Super-resolution methods that use a sequence of images
can further improve the quality. 
However, these methods are beyond the scope of this paper
and only single frame interpolation is considered.

\section{Interpolation problem formulation}
\label{sec:prob}
We consider a setup in which the input LR image to be interpolated 
has been produced from an original HR image through anti-aliasing and decimation.
This way, the LR image does not have evident visual artifacts,
but does have a loss of information.
For instance, the anti-alias low pass filter can be an optical filter in a camera or
a digital filter in an image processing pipeline.

\begin{figure}[t!]
\begin{center}
	\includegraphics[scale=0.36]{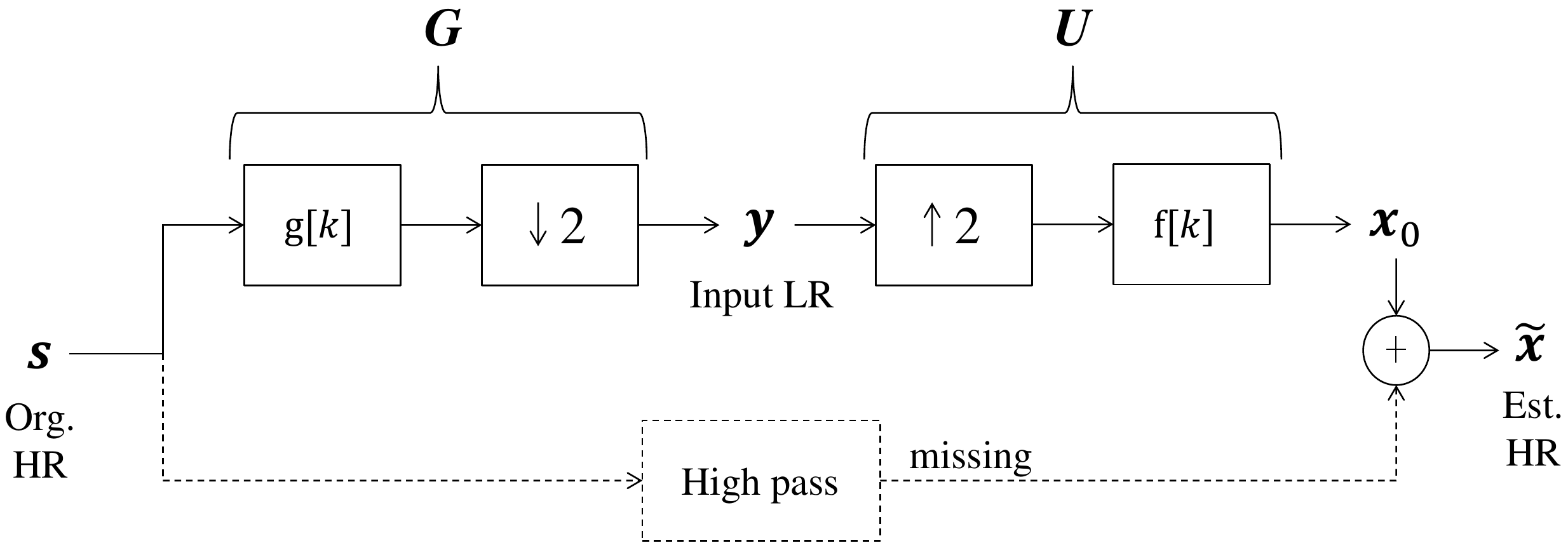}
	\caption[Image recovery problem formulation.]
			{{Image recovery problem formulation.}
			Notation: unknown original HR signal $\vect{s}$; given input LR signal $\vect{y}$; estimated output HR signal $\vect{\tilde{x}}$.
			}
	\label{fig:fb2ch}
  \end{center}
\end{figure}

Let the (unknown) HR original signal of dimensions $N \times 1$ be denoted as $\vect{s}$.
Let the (unknown) low pass filter~$\mathrm{g}[k]$ followed by a decimation together be 
represented as a downsampling matrix $\vect{G}$ of dimension~$n \times N$, where $n < N$.
We are given the result~$\vect{y}$ of dimension~$n \times 1$ as the LR input to the interpolation system,
as depicted in Fig.~\ref{fig:fb2ch}.

One way to estimate an HR signal~$\vect{\tilde{x}}$ is by solving an optimization problem of the form
\begin{equation}
\underset{\vect{\tilde{x}}}{\operatorname{min}} \;\; 
D(\vect{\tilde{x}}) + \lambda \cdot R(\vect{\tilde{x}}).
\end{equation}
where
$D(\vect{\tilde{x}})$ is a fidelity term that penalizes the difference between the given LR signal~$\vect{y}$
and the LR signal obtained by downsampling the estimated HR signal $\vect{\tilde{x}}$ using the downsampler $\vect{G}$, 
while $R(\vect{\tilde{x}})$ is a regularizer that promotes sparsity of the estimated HR signal in a transform domain
and $\lambda$ is a regularization parameter.
Typically, the fidelity term is chosen as an L2 norm, i.e., $D(\vect{\tilde{x}}) = \| \vect{G}\cdot \vect{\tilde{x}} - \vect{y} \|^2$,
which requires the explicit knowledge of~$\vect{G}$.
If we need to find the sparsest solution, we need to choose the penalty function $R(\vect{\tilde{x}})$ as
the L0 (pseudo) norm of the transform coefficients which is unfortunately an NP-hard problem~\cite{Natarajan95}.
If the penalty function $R(\vect{\tilde{x}})$ is chosen to be the L1 norm of the transform coefficients,
it has been shown that it has the effect of promoting sparsity in the transform domain under certain conditions~\cite{Donoho94}.
It then becomes a convex optimization problem and can be solved using general convex solvers, 
e.g., using interior point methods~\cite{Boyd04, Ben-Tal01}.
However, there are simpler gradient-based algorithms for solving functions of this form
and a popular method is called \emph{iterative shrinkage/thresholding algorithm} (ISTA) 
\cite{Figueiredo03, Daubechies04, Zibulevsky10}.
It is also known by other names in signal processing literature, e.g., 
thresholded Landweber method, basis pursuit denoising~\cite{Chen98}, etc.
Optimizing objective functions of this form is an active area of research and 
many fast algorithms, e.g.,~\cite{Beck09}, are being proposed in literature.
Other popular approaches include greedy techniques such as matching pursuits and 
orthogonal matching pursuits~\cite{Mallat93, Tropp04}.

The proposed framework follows the principle 
of image recovery through sparse reconstructions and iterated denoising~\cite{Guleryuz06a, Guleryuz06b}.
This procedure has similarities to ISTA techniques and offers some robustness to noise and transform selection.
While atomic decomposition techniques (L1, greedy, etc) build a solution bottom-up,
iterated denoising takes a top-down approach, 
starting from an initial point and pruning the signal components that it detects as noise.
A detailed comparison of iterated denoising versus atomic decomposition methods for missing data estimation
can be found in~\cite{Guleryuz05}.

\section{Multi-resolution directional transforms}
\label{sec:sr}
%Here, a brief review of fixed transforms suitable for sparse approximation is provided.
One of the main goals of a transform representation is to determine
efficient linear expansions for images. 
Efficiency is generally measured in terms of the number
of elements needed in a linear expansion.
To quantify the number of elements needed for a linear
expansion, image models are employed.
Commonly, images are considered as uniform 2D functions 
separated by singularities (e.g., edges).
The singularities themselves are modeled as smooth curves.
In the past decades, developments in applied harmonic analysis
have provided many useful tools for signal processing.
Wavelets are good at isolating singularities in 1D.
Extending wavelets to 2D,
makes them well adapted to capture point-singularities.
But in natural images, there are mostly line- or curved-
singularities (e.g., directional edges).
These are also known as anisotropic features 
as they are dominant along certain directions.
To capture such features,
there has been extensive study in constructing and 
implementing directional transforms aiming to obtain
sparse representations of such piecewise smooth data. 
The curvelet transform is a directional transform 
which can be shown to provide optimally sparse approximations 
of piecewise smooth images~\cite{CD2004}.  
However, curvelets offer limited localization in the 
spatial domain since they are band limited.  
Also, they are based on rotations which introduce
difficulties in achieving a consistent discrete implementation.
Contourlets are compactly supported directional  
elements constructed based on directional filter banks~\cite{Do2005}.
Directional selectivity in this approach is artificially imposed 
by a special sampling rule of filter banks which 
often causes artifacts. 
Moreover, no theoretical guarantee exists 
for sparse approximations for piecewise 
smooth images.
Recently, a novel directional representation system known as shearlets
has emerged, which provides a unified 
treatment of continuous as well as discrete models, 
allowing optimally sparse representations of 
piecewise smooth images~\cite{KL2011, L2013}. 
This simplified model of natural images, 
which emphasizes anisotropic features, most notably edges, 
is found to be consistent with many models of the 
human visual system~\cite{Kutyniok12}.
%One of the distinctive features of shearlets is that 
%directional selectivity is achieved by shearing in place of rotation.
%Furthermore, shearlets offer a high degree of localization in the 
%spatial domain since they can be compactly supported. 
The framework proposed in this paper is applicable for
all these transforms,
although shearlets is observed to provide the best performance
among the considered transforms.

Multi-resolution directional transforms can also be seen
as filterbanks.
%In other words, the dot product with a transform vector
%is a filtering operation.
The decomposition is implemented using an analysis filter bank, 
while the reconstruction is implemented using a synthesis filter bank.
One branch of the filterbank is designed as a low pass channel 
that captures a coarse representation of the input signal
followed by band- or high-pass channels.
%Fig.~\ref{fig:fb} shows two example filterbanks: 
%Fig.~\ref{fig:mrfb} depicts a multi-resolution filterbank with low-,
%band-, and high-pass components, while,
%Fig.~\ref{fig:drfb} depicts a filterbank that splits the input signal
%into 4 directional bands.
%These two concepts are combined to produce a multi-resolution
%directional filterbank shown in Fig.~\ref{fig:mrdrfb}.
Each of these branches is adapted to capture signal components
at different scales and directions. 
%Such a setup of splitting into bands is also called tiling the time-frequency plane.

\subsection*{Introduction to shearlets}
In modeling image features that are typically anisotropic, 
other than the location and scale,
we would like to include the orientations of the features.
Therefore, a transform is built by combining 
a scaling operator to generate elements at different scales, 
an orthogonal operator to change their orientations, 
and a translation operator to displace these elements over 
the 2D plane~\cite{Kutyniok12}.
Consider a general model for directional transforms built from a generating function~$\psi(t)$
by orienting it using $\vect{O}_s$, scaling it using $\vect{A}_a$, 
and translating it using $\vect{T}_m$,
so that
\begin{equation}
S(\psi) = \vect{T}_m \cdot \vect{A}_a \cdot \vect{O}_s \cdot \psi.
\end{equation}
Below, we discuss the choice of these three operators that 
leads to the so-called shearlet system~$S(\psi)$.

Firstly, to change the orientation of the generating function~$\psi$, 
an obvious choice is a rotation operator. 
However, rotations destroy the integer lattice
(except for trivial rotations that switch the axes).
In other words, 
integer locations may get mapped to fractional locations after a rotation.
This leads to the problem of obtaining a discrete transform that is consistent 
with the continuous transform (where approximation properties have been optimized).
As an alternative orientation operator, consider the \emph{shearing} matrix
\begin{equation}
\vect{O}_s = 
\left[ 
\begin{array}{rr}
1 & s \\
0 & 1 \\
\end{array} 
\right].
\end{equation}
This achieves orientation changes using the slope~$s$ rather than a rotation angle.
It has the advantage of leaving the integer lattice invariant when~$s$ is chosen as an integer.

\pagebreak
Next, the scaling operator is considered. 
Equal scaling along both axes will not be able to capture anisotropic features,
hence different scaling for the axes is required.
Consider the case when one axis is scaled by the factor~$a$ and the other by~$a^{1/2}$,
so that
\begin{equation}
\vect{A}_a = 
\left[ 
\begin{array}{rl}
a & 0 \\
0 & a^{1/2} \\
\end{array} 
\right].
\end{equation}
Although other ratios for scaling the axes are possible, 
this choice, known as parabolic scaling,
optimizes the approximation properties for the 
piecewise smooth image model considered.

Finally, a translation operator is defined that shifts the generating function
\begin{equation}
\vect{T}_m \; \psi(t) \rightarrow \psi(t-m).
\end{equation}

The conditions on the generating function~$\psi$ so that the shearlet system~$S(\psi)$ 
can represent any square-integrable function
are known as admissibility conditions~\cite{Kutyniok12}.

Directional elements capture high frequencies along certain
directions and are not good at representing the low frequencies.
Therefore, in general, a low pass filter is used to extract the low frequency
part and the directional elements are operated on the remaining signal,
leading to the so-called cone-adapted shearlet transform.
By varying the parameters of the shearlet system,
different properties can be achieved, e.g., 
compact support~\cite{Kittipoom12}, orthonormality~\cite{Kutyniok12}, etc.
However, a shearlet system with compact support that
is also orthonormal is, most likely, not achievable~\cite{Houska12}.
Nevertheless, compactly supported shearlet systems have good frame properties, 
i.e., they are close to being a tight frame.

\begin{figure}[t!]
\begin{center}
	\includegraphics[scale=0.48]{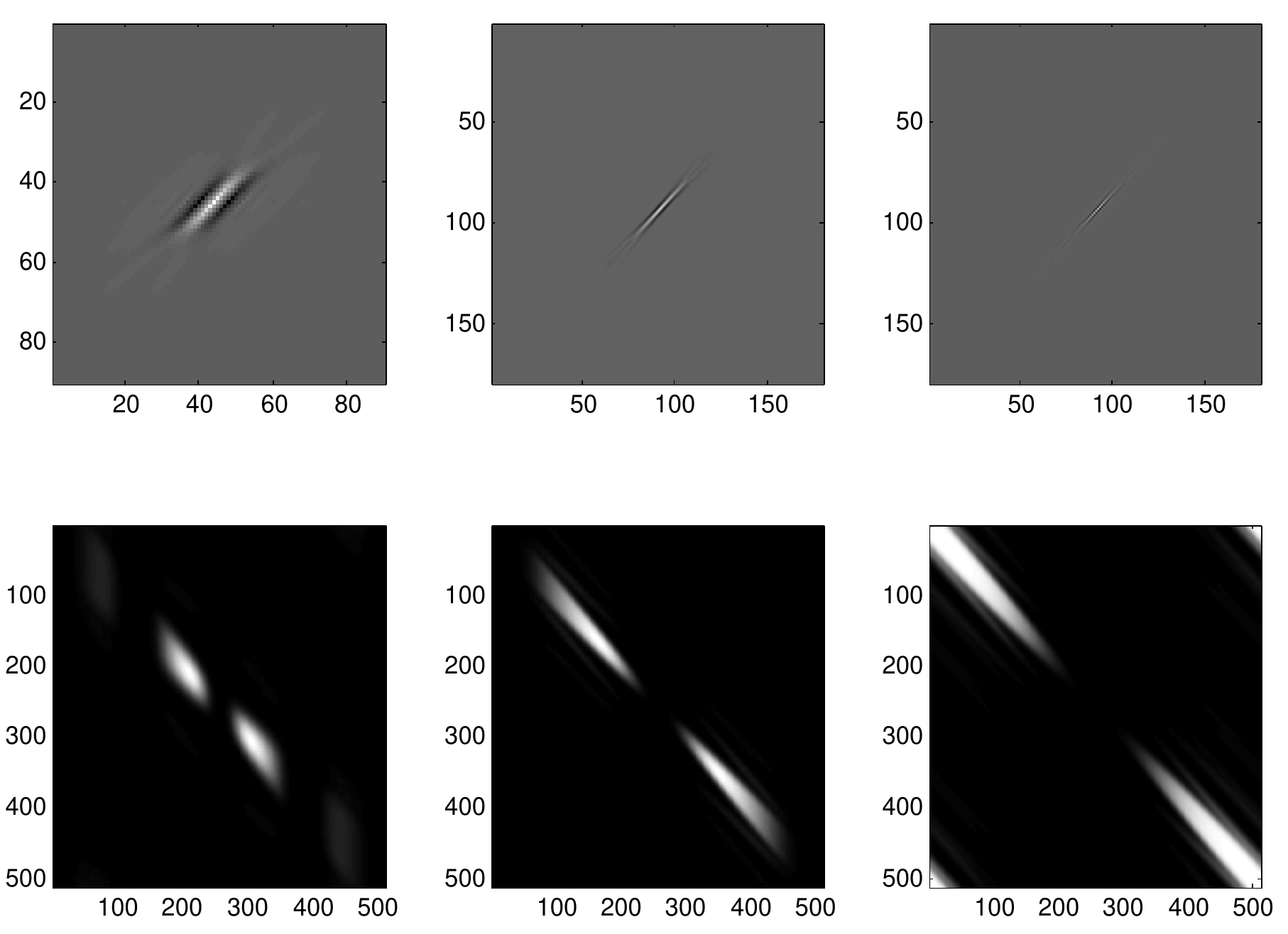}
	\caption{Example of shearlet elements for three scales generated using \cite{Shearlab}
			(top row: spatial domain, bottom row: frequency domain). 
			They are directional and band pass in nature with increasing center frequencies from left to right. \vspace{-3mm}
			}
	\label{fig:sh}
  \end{center}
\end{figure}

Fig.~\ref{fig:sh} shows examples of practical filters (shearlet) 
at a certain orientation and three different scales.

%\begin{figure}[t!]
%\begin{center}
%	\includegraphics[scale=0.6]{figures/mrdrfb}
%	\caption[A multi-resolution directional filterbank]
%			{{A multi-resolution directional filterbank} that splits
%			the 2D input signal into 3 different scales (low pass, band pass, high pass) and 4 directions.
%			}
%	\label{fig:mrdrfb}
%  \end{center}
%\end{figure}

\pagebreak
\section{Proposed Framework for High Frequency Synthesis}
\label{sec:fw}

\begin{figure}[t!]
\begin{center}
	\includegraphics[scale=0.37]{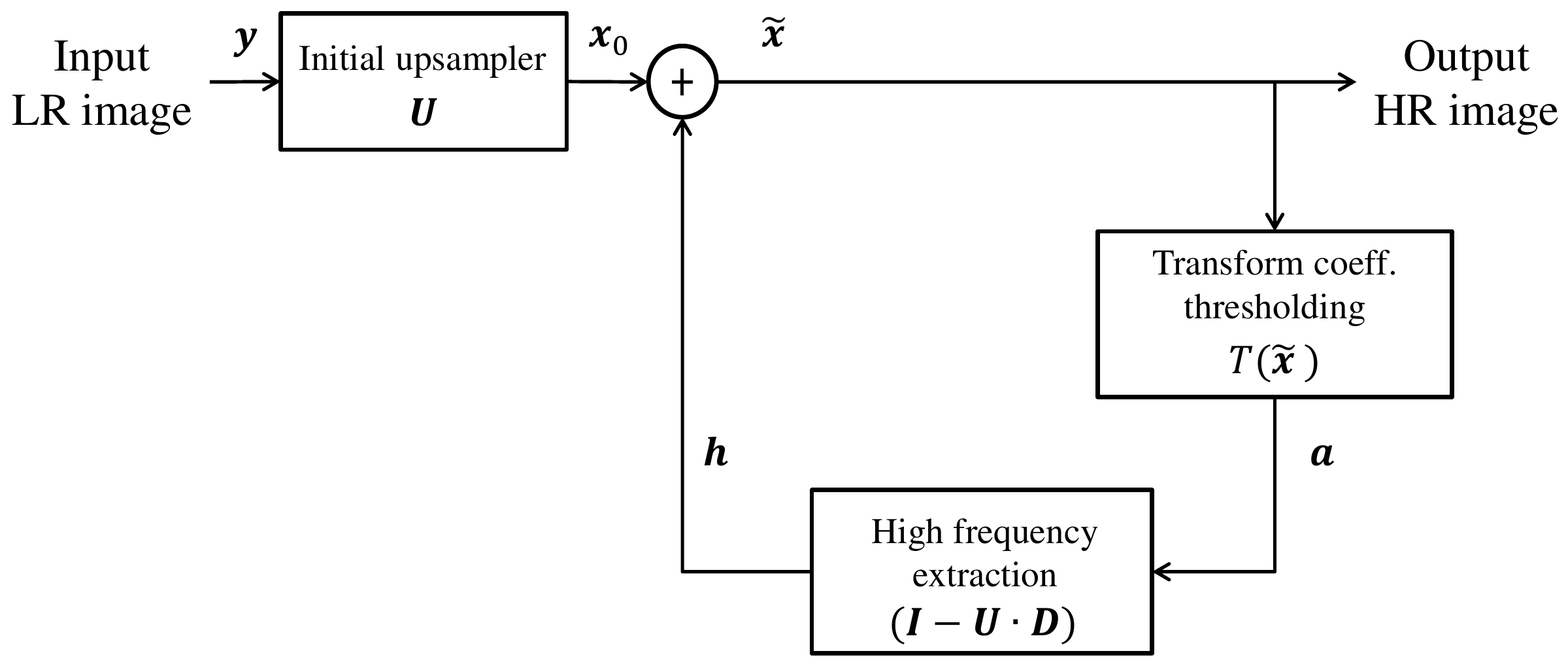}
	\caption[Framework for image interpolation.]
			{{Framework for image interpolation.}
			A linear model, e.g., FIR filter is used to produce an initial upsampled image.
			Then, high frequency components are extracted from a sparse approximation
			and used to refine the initial upsampled image.
			}
	\label{fig:fw}
  \end{center}
\end{figure}

The proposed framework, depicted in Fig.~\ref{fig:fw}, uses the iterated denoising principle. 
It involves:
\begin{itemize} 
\item \textbf{Sparsity constraint:} promoting sparsity, 
e.g., in a multi-resolution directional transform domain 
to improve regularity along edges, and,
\item \textbf{Data constraint:} enforcing constraints according to known data.
\end{itemize}
The problem considered in~\cite{Guleryuz06a} is that of filling missing samples
in an image, where enforcing known data constraints is achieved by replacing input samples
at the known locations after the sparsity promoting step.
However, in the context of image interpolation,
the available LR input image constitutes the known data.
The iterated denoising principle has been applied to 
image interpolation using contourlets in~\cite{Mueller2007},
however, utilizing the knowledge of the LR image generation during the HR image estimation.
Specifically, the LR image was produced through the low pass subband
of a specific wavelet transform and the same transform was used during the 
HR image estimation to enforce the known data constraint.
It is a goal of the proposed approach to interpolate a given LR image without
the knowledge of the exact method generating the LR image.
Therefore, the iterative procedure is redesigned so that the input LR image can be used 
as the known data constraint, 
instead of requiring the low pass subband of a specific wavelet transform.

\subsubsection*{Initial upsampling}
The first stage of the proposed framework 
involves a conventional FIR filter based interpolation of the LR signal~$\vect{y} \in \mathbb{R}^n$ 
to produce an initial HR estimate $\vect{x}_0 \in \mathbb{R}^N$.
It can be expressed in a vector notation as
\begin{equation}
\vect{x}_0 = \vect{U} \cdot \vect{y},
\label{eq:initup}
\end{equation}
where the rows of the upsampler~$\vect{U}$ specify
the filter coefficients used to generate the samples of~$\vect{x}_0$.
This process can also be seen as a zero insertion in the spatial domain
followed by a low pass filter to remove the spectral replication due to the zero insertion.
Since the coefficients in~$\vect{U}$ act as a low pass filter,
some high pass details would be missing/distorted in the initial HR estimate
compared to the HR original.
Therefore, the initial HR estimate is seen as a noisy version of an unknown 
HR original and then refined in an iterative manner.
The refined HR signal is denoted as~$\vect{\tilde{x}}$, which, 
during the first iteration, is set as~$\vect{\tilde{x}}_1 = \vect{x}_0$.

\subsubsection*{Sparsity promoting}
As stated earlier, a dictionary consisting of multi-resolution directional transform 
elements is considered.
Promoting sparsity in such a dictionary results in regular directional structures
in the approximated signal.
Denoting the iteration number of refinement as~$k$, the sparsity promoting step operates as follows:
\begin{itemize}
\item the signal $\vect{\tilde{x}}_k$ is forward transformed to the selected domain
(resulting in directional components in different scales),
\item the transform coefficients are hard-thresholded, and
\item inverse transformed to generate an approximation~$\vect{a}_k$.
\end{itemize}
The overall operation is written compactly as, $\vect{a}_k = T(\vect{\tilde{x}}_k)$.
This denoising step is closely related to techniques such as ISTA for L1 regularization 
but has some differences~\cite{Guleryuz05}.

\subsubsection*{Known data constraint}
Then, we enforce the known LR data constraint.
It is done by assuming that the initial upsampled signal~$\vect{x}_0$
is equal to the low pass channel of a two-channel filterbank, depicted in Fig.~\ref{fig:fb2ch}.
The missing high pass channel is generated by using the approximated signal~$\vect{a}_k$.
Hence, it is required to separate the signal~$\vect{a}_k$ into low pass and high pass channels.
At this stage we face the issue of the unknown downsampler that generated the input LR signal~$\vect{y}$.
A blind deconvolution would be necessary to jointly estimate the unknown downsampler and undo its effect,
which is very difficult.
Instead, a downsampler~$\vect{D}$ is chosen so that the product $\vect{P} = \vect{U} \cdot \vect{D}$ acts as a projection matrix,
i.e., $\vect{P}^2 = \vect{P}$.
Then, enforcing the known data constraint can be implemented by only considering the components of~$\vect{a}_k$
that do not fall on the low pass projection space, i.e., using the high pass components of~$\vect{a}_k$ for refinement.
However, there could be a mismatch between the utilized~$\vect{D}$ and the actual external downsampler that
produced the LR signal.
This will be experimentally studied in Sec.~\ref{sec:dsinf} 
by fixing the upsampler and downsampler of the proposed system,
but varying the actual external downsampler to produce different LR inputs to the proposed system
and recording the performance variation.

Summarizing, we can write the low pass $\vect{l}_k$ and the high pass $\vect{h}_k$ decomposition of the approximated signal
$\vect{a}_k$ as
\begin{eqnarray}
\vect{l}_k  &=& \vect{U} \cdot \vect{D} \cdot\vect{a}_k,  \nonumber \\
\vect{h}_k &=& (\vect{I} - \vect{U}\cdot\vect{D}) \cdot \vect{a}_k.
\end{eqnarray}

\subsubsection*{Refinement step}
The high pass component~$\vect{h}_k$ is used for refinement
by adding it to the initial HR estimate~$\vect{x}_0$, 
to produce a refined HR estimate~$\vect{\tilde{x}}_{k+1}$, i.e.,
\begin{equation}
\vect{\tilde{x}}_{k+1} \longleftarrow \vect{x}_0 + \vect{h}_k.
\label{eq:rf}
\end{equation}
For the first iteration, the vector~$\vect{h}_0$ is initialized to zero,
therefore, $\vect{\tilde{x}}_1 = \vect{x}_0$.
%The overall expression for each iteration can be written as,
%$\vect{y}^{(i+1)} =\vect{U} \cdot \vect{x} + (\vect{I}-\vect{P}) \cdot \vect{a}^{(i)}.$
%Since $\vect{y}^{(0)}=\vect{U} \cdot \vect{x}$ is already in the column space of $\vect{U}$, 
%it is unaltered by a projection step, i.e., $\vect{y}^{(0)} = \vect{P} \cdot \vect{y}^{(0)}$.

By combining Eq.~\ref{eq:initup} through Eq.~\ref{eq:rf}, 
the overall system connecting the input LR signal~$\vect{y} \in \mathbb{R}^n$
to the refined HR signal~$\vect{\tilde{x}}_{k+1} \in \mathbb{R}^N$ can now be expressed as
\begin{equation}
\vect{\tilde{x}}_{k+1} \longleftarrow \vect{U} \cdot \vect{y} + (\vect{I}-\vect{U} \cdot \vect{D}) \cdot T(\vect{\tilde{x}}_k).
\label{eq:itd}
\end{equation}

\begin{figure}[t!]
\begin{center}
	\includegraphics[scale=0.6]{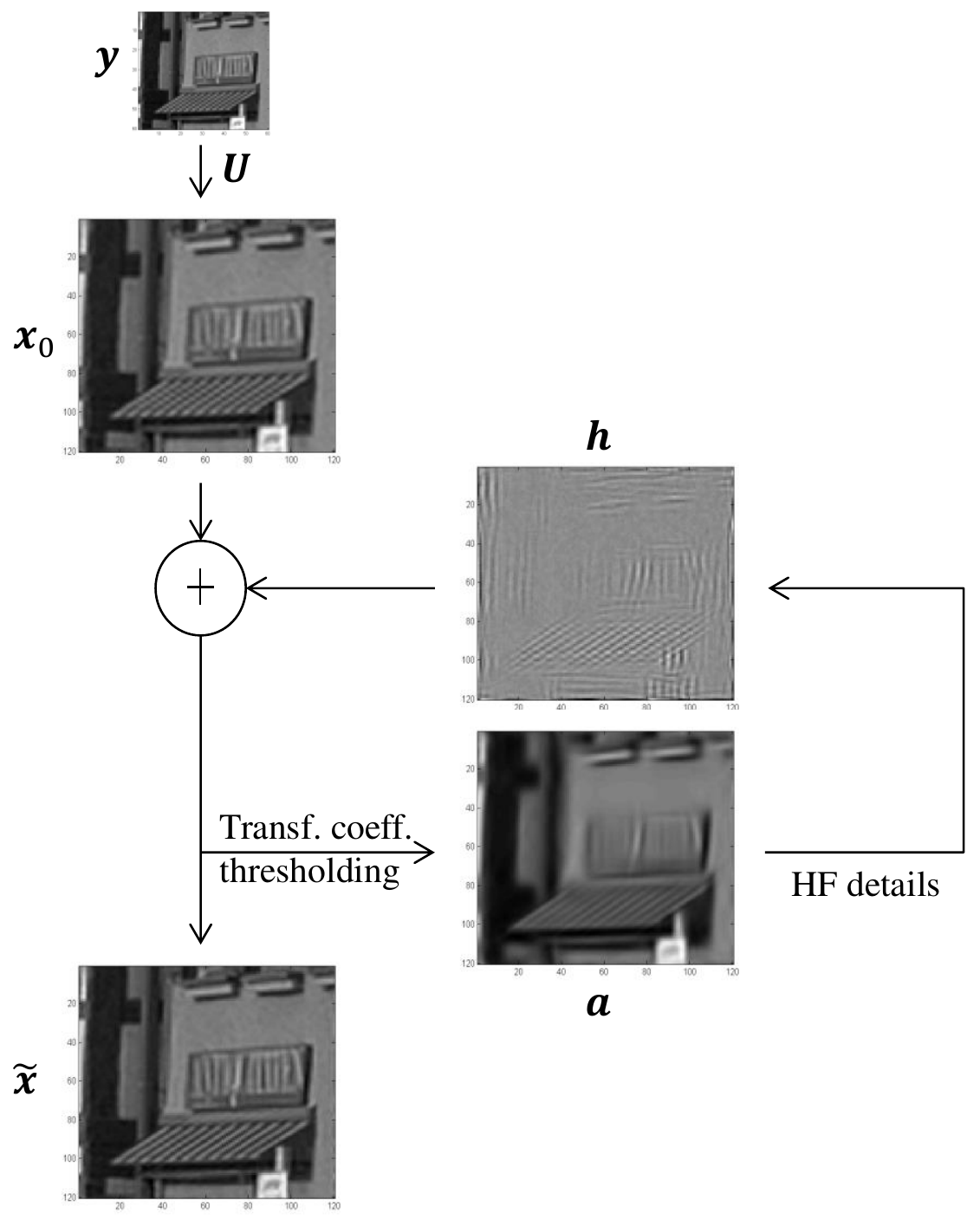}
	\caption[Example images at each stage of processing.]
			{{Example images at each stage of processing.} 
			Figure shows the quality of the initial HR estimate, the result of transform domain thresholding and the estimated high pass details.
			Notice that the diagonal lines become slightly sharper after adding the estimated details. 
			\vspace{-4mm}
			}
	\label{fig:iter}
  \end{center}
\end{figure}

The iterative procedure is repeated for a certain maximum number of iterations
and $\vect{\tilde{x}}_{k+1}$ after the last iteration is taken as the output HR image.

Fig.~\ref{fig:iter} depicts example images during different stages of the proposed approach.
It can be seen that the initial upsampled image is blurry around the diagonal edges.
The step of transform domain thresholding retains only the dominant information.
After adding the high frequency part, the resulting image looks slightly sharper.

\section{Simulation results}
\label{sec:results}

The proposed algorithm is tested for both subjective and objective performance.
For a subjective evaluation, original images are directly used as LR inputs
and the HR outputs are inspected for visual quality/artifacts.
Using original images as LR inputs avoids downsampling artifacts in inputs. 
However, for an objective evaluation, we require a reference HR image.
To this end, a 11-tap FIR anti-alias filter, 
that is tested in the ITU-T/ISO-IEC evaluations of Scalable Video Coding~\cite{Dong2012},
is used before decimation to generate an LR image and the original image is used as the reference HR image
to measure the PSNR.
The coefficients of the 11-tap filter for 2x downsampling are
\small $[2, -2, -9, 3, 40, 60, 40, 3, -9, -2, 2]/128$. \normalsize
In all the experiments, this filter remains unknown to the proposed interpolation system.
Additionally, in Sec.~\ref{sec:dsinf}, the proposed system is kept fixed and the external downsamplers are varied
to record the performance variation.

There are many free parameters to be chosen in the proposed method, 
such as the initial upsampling filter, number of scales and directions in the transform, 
thresholds levels for hard thresholding in the transform domain, etc.
A joint optimization of all these internal parameters involves a large search space.
Hence, a simpler approach is followed here, 
where we first select an initial set of parameters
and optimize some free parameters keeping the others fixed,
for 2x upsampling.
The optimization of free parameters is conducted using a training set (16 images)
and the final performance is evaluated on a test set (200 images). 
The training and test sets are disjoint.
Throughout the optimization, the proposed method with the chosen parameter set is compared to a
system with an 8-tap FIR filter without any iterative refinement to record the average PSNR gain in the training set.
Although a 12-tap filter provides a higher PSNR, it is not preferred as a reference, 
since some ringing artifacts can be noticed in the 12-tap filter results.
\vspace{3mm}
\subsubsection*{Initial upsampler and Downsampler for high frequency extraction}
In the first stage of the proposed framework, 
the input LR image is upsampled using~$\vect{U}$.
The rows of~$\vect{U}$ are filled with FIR filter coefficients so that 
the samples in the HR grid corresponding to zero phase shift in the LR grid are copied directly 
and the required fractional shifts are produced using FIR filters.
To this end, for 2x upsampling, five different filters are considered
which are given in Tab.~\ref{tab:initups}.

\begin{table}[t!]
\begin{center}
\footnotesize
{
\begin{tabular}{|c|c|}
\hline
\textbf{Symbol} & \textbf{Interpolation filter coefficients} \\ 
\hline
u2 & $[1, 1]/2$ \\
u4 & $[-1, 9, 9, -1]/16$ \\
u6 & $[1, -5, 20, 20, -5, 1]/32$ \\
u8 & $[-1, 4, -11, 40, 40, -11, 4, -1]/64$ \\
u12 & $[-1, 4, -10, 22, -48, 161, 161, -48, 22, -10, 4, -1]/256$ \\
\hline
\end{tabular}
}
\end{center}
\caption[Set of FIR filters considered for initial interpolation]
{Set of FIR filters considered for initial interpolation.}
\label{tab:initups}
\end{table}

%\subsubsection*{Downsampler for high frequency extraction}
Next, a downsampler~$\vect{D}$ is designed to enforce the known data constraint.
Ideally, a sinc filter for~$\vect{U}$ and~$\vect{D}$ results in~$\vect{P}=\vect{U}\cdot\vect{D}$ 
being a projection operator~\cite{Strang2009}.
However, it will be shown in Sec.~\ref{sec:initupsdown} that FIR filter approximations in~$\vect{U}$ and~$\vect{D}$ 
are sufficient for the purpose of high frequency extraction in the current setup.
To this end, five different anti-alias filters are evaluated for 2x downsampling, 
given in Tab.~\ref{tab:dsfilters}.
All the considered filters are odd-length and symmetric, hence they do not induce any phase shift.

\begin{table}[t!]
\begin{center}
\footnotesize
{
\begin{tabular}{|c|c|c|}
\hline
\textbf{Symbol} & \textbf{N-tap} & \textbf{Anti-aliasing filter coefficients} \\ 
\hline
d3 & 3  & $[1, 2, \cdots]/4$ \\
d9 & 9  & $[-1, 0, 9, 16, \cdots]/32$ \\
d13 & 13  & $[1, 0, -5, 0, 20, 32, \cdots]/64$ \\
d17 & 17  & $[-1, 0, 4, 0, -11, 40, 64, \cdots]/64$ \\
d25 & 25 & $[-1, 0, 4, 0, -10, 0, 22, 0, -48, 0, 161, 256, \cdots]/256$ \\
\hline
\end{tabular}
}
\end{center}
\caption[Set of FIR filters considered for anti-aliasing in high frequency extraction]
{Set of FIR filters considered for anti-aliasing in high frequency extraction.
Dots denote repetition of coefficients with mirror symmetry.
\vspace{-4mm}
}
\label{tab:dsfilters}
\end{table}

\vspace{3mm}
\subsubsection*{Directional transform parameters}
A compactly supported shearlet transform~\cite{Shearlab, L2013} is chosen for the 
multi-resolution directional representation.
The initial configuration used for the shearlet transform is:
1 low pass component, $2^3$ directional band pass components 
and $2^3$ directional high pass components.
These settings can be compactly represented in an array as $[0, 3, 3]$,
where the entries of the array are interpreted as exponents of two.
The number of entries in the array denotes the number of scales used. 
For instance, $[0, a]$ represents a configuration consisting of two scales:
one low pass component and $2^a$ directional high pass components. 
The configuration $[0, a, b]$ represents three scales: 
one low pass component, $2^a$ directional band pass components, 
and $2^b$ directional high pass components.
The computation of shearlet transform coefficients and 
the reconstruction are carried out as multiplications in the Fourier domain 
instead of convolutions in the spatial domain to reduce the computational complexity.
The stages of sparsity enforcement and high frequency extraction are repeated 8 times.
The threshold value for hard-thresholding the shearlet coefficients is set to 100 
and decreased by a multiplicative factor of 0.6 in each iteration.
The proposed framework is also tested with the contourlet transform.
For a direct comparison of the contourlet and shearlet dictionaries,
the upsampling and downsampling filters in the proposed framework are kept fixed 
and only the dictionaries are switched. 
The threshold values for the contourlet case are taken from~\cite{Mueller2007}.

\subsection{Influence of initial interpolator \& high frequency extractor}
\label{sec:initupsdown}
\begin{figure}[t!]
\begin{center}
	\includegraphics[scale=0.55]{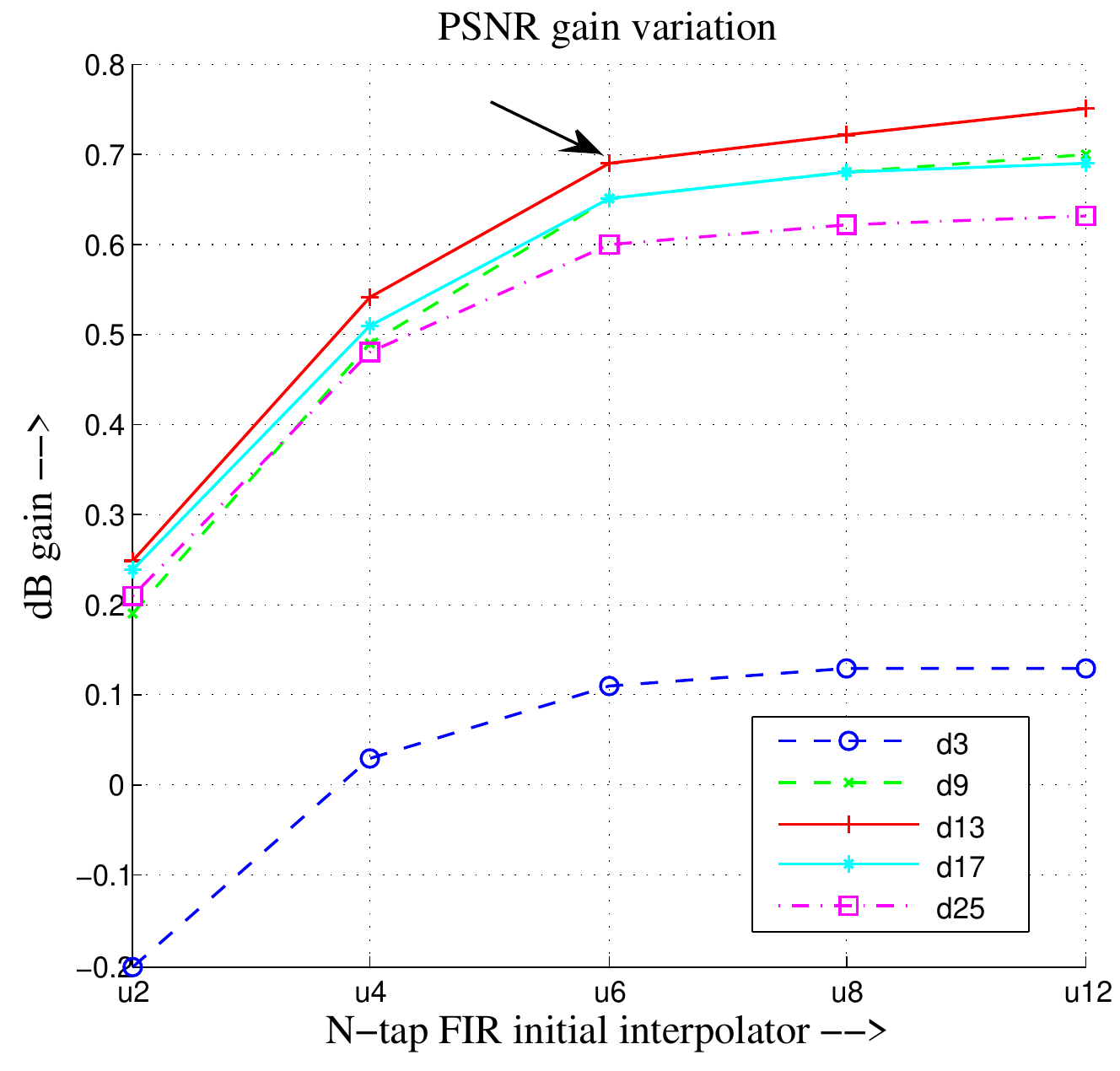}
	\caption[Influence of initial upsampler; and downsampler for HF extraction.]
			{{Influence of initial upsampler; and downsampler for HF extraction.}
			Average PSNR difference (dB) between reference (8-tap FIR interpolator)
			and test (proposed refinement approach with different combinations of $\vect{U}$ and $\vect{D}$) 
			for a dataset of 16 training images.
			PSNR improvements for initial interpolation filters beyond 6-tap are rather small. 
			\vspace{-4mm}
			}
	\label{fig:initupsdown}
  \end{center}
\end{figure}

The influence of~$\vect{U}$ and~$\vect{D}$ on the final HR result is studied here.
To this end, each interpolation filter from the set \{u2, u4, ..., u12\} is combined with
a downsampling filter from the set \{d3, d9, ..., d25\} and 25 HR results are produced
for each LR input,
i.e., the entire product space is tested.
Fig.~\ref{fig:initupsdown} shows the test results for each tested parameter combination,
in the form of average PSNR difference to the 8-tap FIR (u8) reference system.
In the y-axis, the 0~dB gain level represents a PSNR that is the same as the reference system. 
It can be seen that the 3-tap anti-alias filter d3 is not well suited for the system,
because it leaves too much aliasing.
The remaining anti-alias filters from the set perform relatively well.
The best PSNR performance is observed when the 13-tap anti-alias filter is combined with a 12-tap interpolator,
giving 0.75~dB gain over the reference 8-tap FIR interpolator.
However, PSNR improvements for interpolation filters beyond 6-tap are rather small and 
the 12-tap interpolation filter might introduce ringing artifacts in the initial upsampled image.
Therefore, the combination of the 6-tap interpolation filter and the 13-tap downsampling filter is chosen for further investigation.

\subsection{Selection of the number of scales and directions in transform}
\begin{figure}[t!]
\begin{center}
	\includegraphics[scale=0.55]{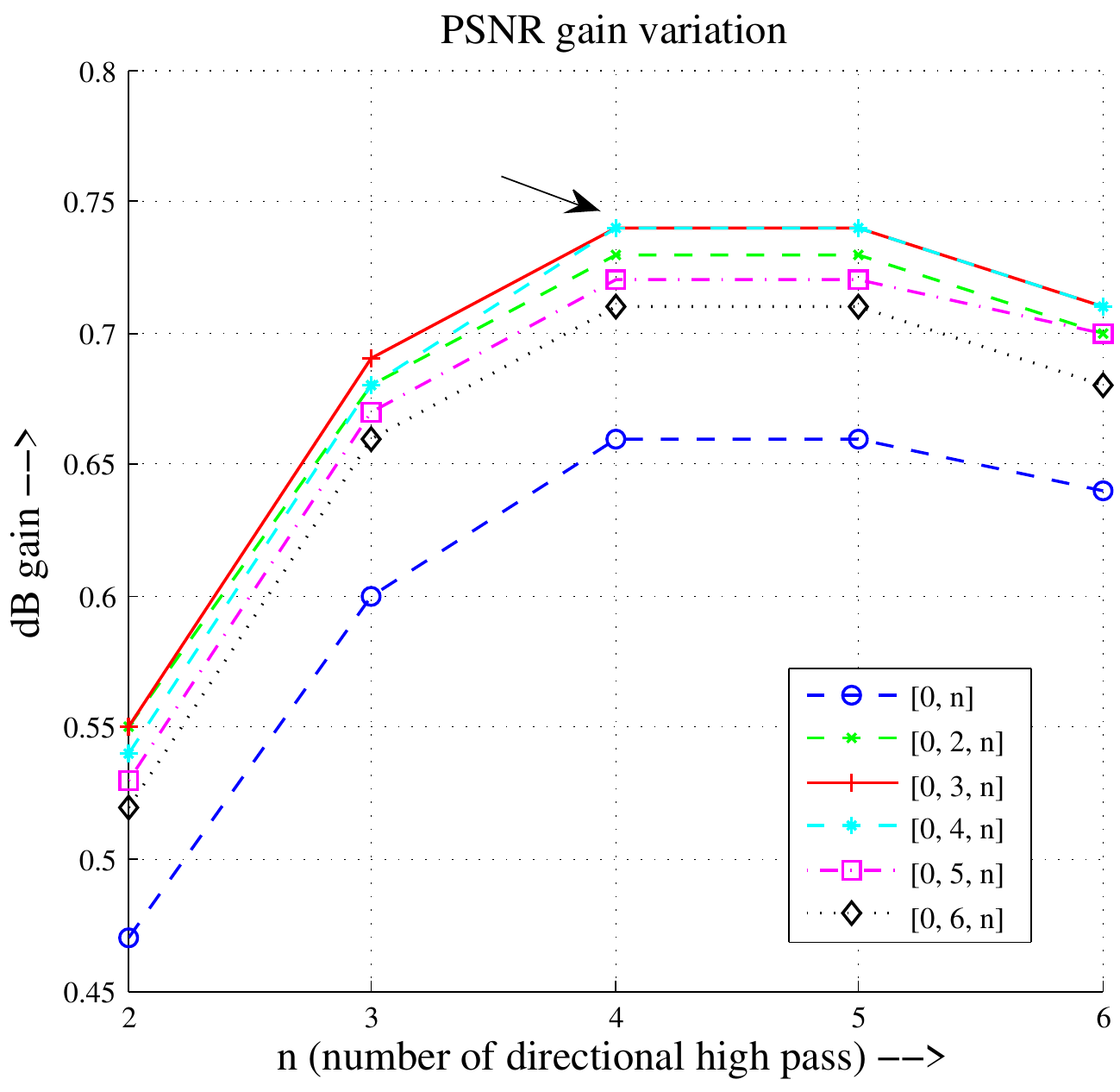}
	\caption[Influence of the number of scales and directions.]
			{{Influence of the number of scales and directions.}
			Each tested configuration is represented in the array notation introduced.
			The configuration [0, 3, 4], i.e., splitting the signal into one low pass,
			$2^3$ directional band pass and $2^4$ directional high pass components, 
			is observed to give the best results.
			\vspace{-4mm}
			}
	\label{fig:numscldir}
  \end{center}
\end{figure}

Next, the influence of the number of scales and directions for 
thresholding the estimated HR image in the transform domain is studied.
The tested configurations are compactly represented in the same array format
described earlier.
PSNR results using the proposed system in the tested configurations are 
compared to the reference 8-tap FIR (u8) system and
the observed average gains are shown in Fig.~\ref{fig:numscldir}.
It can be seen that the configuration $[0, 3, 4]$, i.e., 
one low pass, 8 directional band pass and 16 directional high pass components, 
provides the best performance among the tested transforms (0.74 dB improvement over reference). 

In fact, for a 2x upsampling, we expect that only around half the frequency components need refinement,
for which, using two scales should be sufficient.
However, it can be seen from Fig.~\ref{fig:numscldir} that the three scale configurations, 
namely, $[0, 2, \mathrm{n}], [0, 3, \mathrm{n}], \cdots, [0, 6, \mathrm{n}]$, 
perform better than the two scale configuration $[0, \mathrm{n}]$.
It suggests that an intermediate scale provides a soft transition from low- to high- frequency components
for refinement.
In other experiments (not shown in figure), it is observed that using more than 
three scales for 2x upsampling does not increase the gain further.

\subsection{Threshold selection for sparse approximation}
\begin{figure}[h!]
\begin{center}
	\includegraphics[scale=0.55]{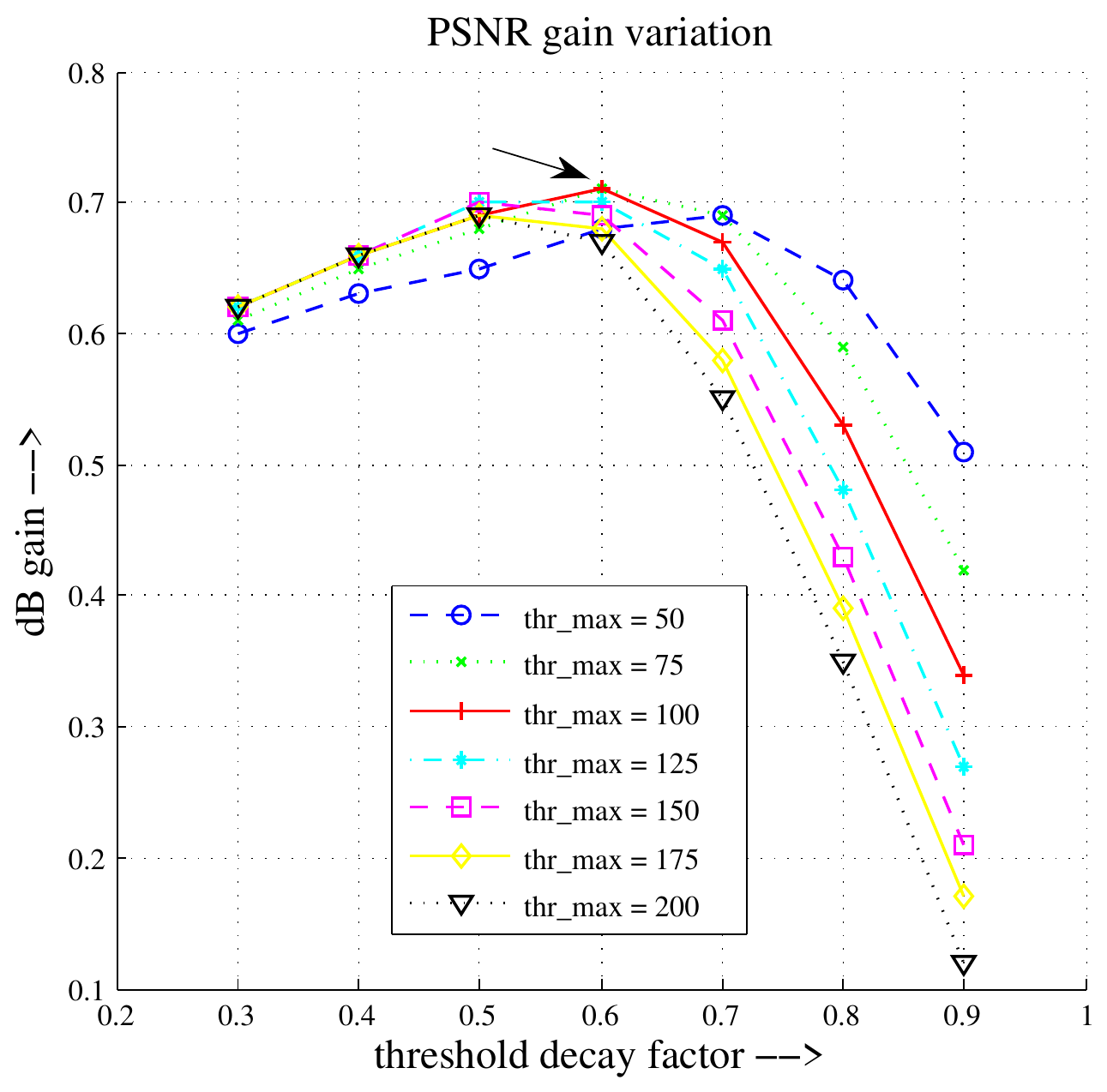}
	\caption[Threshold selection experiments.]
			{{Threshold selection experiments.}
			The threshold for the first refinement iteration is denoted as thr\_max
			and decreased exponentially in each iteration by a decay factor (x-axis).
			The PSNR gain compared to an 8-tap FIR interpolator is recorded (y-axis).
			The system shows best gains for thr\_max between 75 to 125
			with a decay factor around 0.5 to 0.6.
			\vspace{-4mm}
			}
	\label{fig:thres}
  \end{center}
\end{figure}

The effect of thresholding in the shearlet domain on the final interpolation quality
is hard to express analytically.
To this end, two parameters for heuristic optimization are identified:
(a) threshold for the first iteration of refinement, denoted as thr\_max, and
(b) a multiplicative decay factor to decrease the threshold in each iteration.
The maximum number of iterations is set as 8 to limit the overall computational complexity.
For instance, thr\_max = 200 and decay = 0.7 generates the following
thresholds: $\{200, 200 \times 0.7, 200 \times 0.7^2, \cdots, 200 \times 0.7^7 \}$.
The low pass components of the shearlet transform are not thresholded
and the same threshold value is used for the remaining components,
although a band wise optimization of thresholds may further improve the performance.
PSNR results of the proposed method with chosen parameters are 
compared to the reference 8-tap FIR (u8) system and the PSNR gain is computed.
Average PSNR gains on the training set is plotted in Fig.~\ref{fig:thres}.
It can be observed that thr\_max = 75, 100 and 125 perform well
with a decay factor of 0.5 or 0.6.
The combination of thr\_max = 100 and decay = 0.6,
which is the same as our initial setting,
is selected for the final evaluation on the test set.

\begin{table*}[htb]
\begin{center}
\small
{
\begin{tabular}{|c|c|}
\hline
\textbf{External downsampler to produce LR input} & \textbf{Proposed vs. 8-tap FIR} \\ 
\hline
$[-1, 0, 9, 16, 9, 0, -1]/32$ & 0.66 dB\\
$[-2, 0, 64, 132, 64, 0, -2]/256$ & 0.58 dB\\
$[1, 0, -5, 0, 20, 32, 20, 0, -5, 0, 1]/64$ & 0.67 dB\\
$[1, 0, -11, 0, 74, 128, 74, 0, -11, 0, 1]/256$ & 0.66 dB\\
$[-1, 0, 4, 0, -17, 0, 78, 128, 78, 0, -17, 0, 4, 0, -1]/256$ & 0.66 dB\\
$[1, 0, -2, 0, 7, 0, -21, 0, 79, 128, 79, 0, -21, 0, 7, 0, -2, 0, 1]/256$ & 0.60 dB\\
\hline
\end{tabular}
}
\end{center}
\caption[Different downsamplers to generate input LR]
{{Influence of using different downsampling filters to generate LR images.}
For each HR image, six different LR images are generated using 2x downsampling filters given in the first column.
It can be seen that the proposed method achieves stable results and the external downsampling filter does not greatly influence the gains.
}
\label{tab:ds}
\end{table*}

\begin{table*}[htb]
\begin{center}
\small
{
\begin{tabular}{|l|c|c|c|c|c||c|c|}
\hline
\textbf{Image name} & \textbf{Bicubic} & \textbf{Directional} & \textbf{Cubic spline} & \textbf{8-tap} & \textbf{12-tap} & \textbf{Contourlet} & \textbf{Shearlet} \\ 
\hline
bikes& 26.68& 26.20& 27.02& 27.23& 27.32& 27.63& 28.38\\
building2& 23.83& 22.89& 24.08& 24.28& 24.34& 24.58& 24.84\\
buildings& 23.85& 23.32& 24.06& 24.23& 24.29& 24.51& 24.78\\
caps& 35.60& 35.38& 35.78& 36.06& 36.13& 36.33& 37.03\\
coinsinfountain& 30.56& 29.60& 30.44& 31.08& 31.16& 31.62& 32.08\\
flowersonih35& 23.74& 22.76& 23.87& 24.13& 24.19& 24.47& 24.71\\
house& 31.09& 30.62& 31.38& 31.52& 31.60& 31.73& 32.14\\
lighthouse2& 29.19& 28.55& 29.44& 29.55& 29.61& 29.78& 30.07\\
monarch& 31.87& 31.04& 32.37& 32.59& 32.71& 33.03& 33.85\\
ocean& 32.17& 31.70& 32.23& 32.47& 32.52& 32.62& 32.93\\
paintedhouse& 28.23& 27.64& 28.50& 28.65& 28.71& 28.90& 29.35\\
parrots& 34.82& 34.39& 35.36& 35.59& 35.70& 35.88& 36.59\\
plane& 31.47& 30.32& 31.59& 31.86& 31.92& 32.30& 32.78\\
rapids& 29.42& 28.73& 29.67& 29.91& 29.98& 30.18& 30.66\\
sailing1& 28.60& 27.77& 28.81& 28.92& 28.97& 29.14& 29.34\\
stream& 24.73& 24.03& 24.93& 25.08& 25.14& 25.29& 25.50\\
\hline
Average (Train) & 29.12 & 28.43 & 29.35 & 29.57 & 29.64 & 29.87 & 30.32\\
PSNR diff. (Train) &-1.20 & -1.88 & -0.97 & -0.74 & -0.67 & -0.44 & - \\
\hline
\textbf{PSNR diff. (Test)} & \textbf{-1.09} & \textbf{-1.86} & \textbf{-0.81} & \textbf{-0.63} & \textbf{-0.56} & \textbf{-0.47} & \textbf{-} \\
\hline
\end{tabular}
}
\end{center}
\caption[PSNR results in dB for 2x interpolation]
{{PSNR results in dB for 2x interpolation comparing seven methods.}
Three linear approaches (bicubic, cubic spline, and 8-tap FIR) and 
two non-linear approaches (Directional~\cite{Li2001} and contourlet~\cite{Do2005}) 
are compared to the proposed technique.
The PSNR difference over 16 training and 200 test images are summarized.
\vspace{-3mm}
}
\label{tab:res}
\end{table*}

\subsection{Influence of external downsamplers to generate LR images}
\label{sec:dsinf}
With the system parameters fixed, 
the influence of the external downsampling filter used to generate an LR input from the HR original is studied in this experiment.
To this end, six different downsampling filters (approximately halfband cut-off) are used and six LR images are generated for each HR original.
The test is conducted such that the proposed method remains fixed and is unaware of the actual external downsampler that 
has been used to generate the LR input.
As a reference, the 8-tap FIR filter (u8) is used to interpolate the same LR image and the resulting PSNR is measured.
Then, the PSNR difference to the reference result is recorded.
The average PSNR gain on the training set is summarized in Tab.~\ref{tab:ds}.
It can be seen from the result that the gains from the proposed technique do not vary much when changing
the downsampling filters, as long as there is not much aliasing in the generated LR images.

\subsection{Final results on training and test set}
The performance of the proposed method is compared to various 
linear and non-linear methods.  
Among linear methods: bicubic interpolation (u4),  
8-tap filter (u8), 12-tap filter (u12) and cubic spline interpolation
are considered.
The cubic spline approach is implemented as an IIR prefilter 
to compute spline coefficients followed by a 4-tap FIR filter for interpolation.
Among the non-linear models, 
a directional interpolation (NEDI~\cite{Li2001}) technique is considered.
The proposed framework is tested with contourlet and shearlet transforms.
The parameters for the contourlet case are taken from~\cite{Mueller2007}.

The objective performance numbers of the overall system with the selected
parameter settings are summarized in Tab.~\ref{tab:res} for the training and test set.
As can be seen, the proposed approach consistently achieves a higher PSNR 
compared to the other methods tested.
On an average, a PSNR improvement of 0.74~dB is achieved compared to the 8-tap
filter for the considered training images.
As a test set,
200 images from the Berkeley Segmentation Dataset~\cite{Martin2001}
are used.
Average PSNR improvements are recorded in the last row of Tab.~\ref{tab:res}.
Compared to the 8-tap FIR filter, an average gain of about 0.63~dB is observed.
The maximum gain and the minimum gain in the test set, compared to the 8-tap filter, 
are observed to be 3.13~dB and 0.14~dB, respectively.
The average gains observed on the test set are close to the training set numbers.

\subsection*{Subjective evaluation}
\begin{figure*}[t]
	\centering
	\subfloat[]
	{
		\label{fig:f1}
		\includegraphics[scale=0.5]{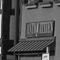}
	}            
	\subfloat[]
	{
		\label{fig:f2}
		\includegraphics[scale=0.5]{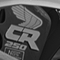}
	}\\  
	\subfloat[Directional~\cite{Li2001}]
	{
		\label{fig:f3}
		\includegraphics[scale=0.59]{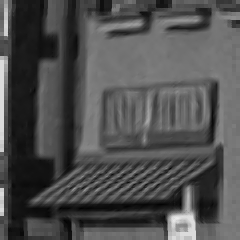}
	}
	\subfloat[Cubic spline]
	{
		\label{fig:f4}
		\includegraphics[scale=0.59]{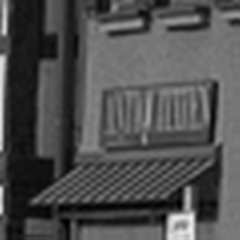}
	}
	\subfloat[Proposed]
	{
		\label{fig:f5}
		\includegraphics[scale=0.59]{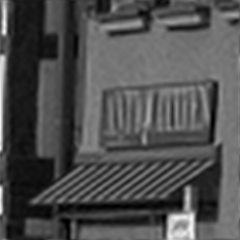}
	}\\
	\subfloat[Directional~\cite{Li2001}]
	{
		\label{fig:f6}
		\includegraphics[scale=0.59]{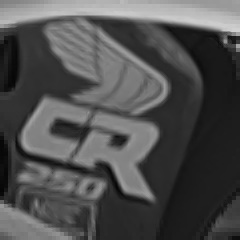}
	}
	\subfloat[Cubic spline]
	{
		\label{fig:f7}
		\includegraphics[scale=0.59]{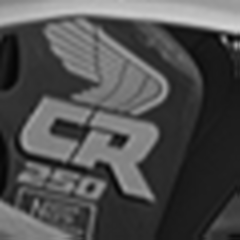}
	}
	\subfloat[Proposed]
	{
		\label{fig:f8}
		\includegraphics[scale=0.59]{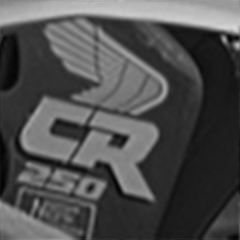}
	}\\
	\caption[Example 4x interpolation results.]
			{{Example 4x interpolation results.} 
			Input patches of size $64\times64$ in (a) and (b) are upsampled to $256\times256$. 
			In (c)~the diagonal stripes show jaggedness,
			in (d)~the diagonal stripes are blurred.
			In (f)~some artifacts can be noticed,
			in (g)~the numbers and the rectangular frame below are blurry.
			The results of the proposed approach, (e) and (h), appear slightly sharper without evident artifacts.
			\vspace{-2mm}
			}
	\label{fig:upsres1}
\end{figure*}

\begin{figure*}[t]
	\centering
	\subfloat[]
	{
		\label{fig:t1}
		\includegraphics[scale=0.5]{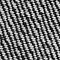}
	}            
	\subfloat[]
	{
		\label{fig:t2}
		\includegraphics[scale=0.5]{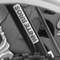}
	}\\  
	\subfloat[Directional~\cite{Li2001}]
	{
		\label{fig:t3}
		\includegraphics[scale=0.59]{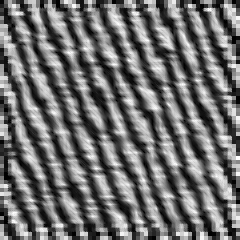}
	}
	\subfloat[Cubic spline]
	{
		\label{fig:t4}
		\includegraphics[scale=0.59]{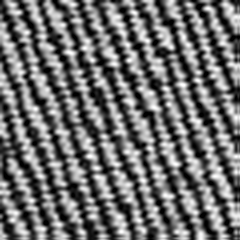}
	}
	\subfloat[Proposed]
	{
		\label{fig:t5}
		\includegraphics[scale=0.59]{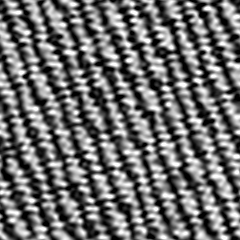}
	}\\
	\subfloat[Directional~\cite{Li2001}]
	{
		\label{fig:t6}
		\includegraphics[scale=0.59]{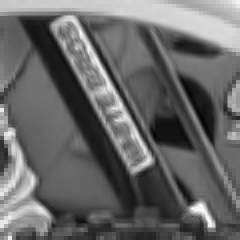}
	}
	\subfloat[Cubic spline]
	{
		\label{fig:t7}
		\includegraphics[scale=0.59]{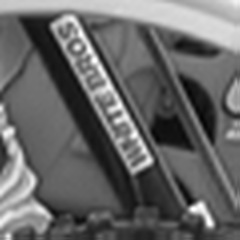}
	}
	\subfloat[Proposed]
	{
		\label{fig:t8}
		\includegraphics[scale=0.59]{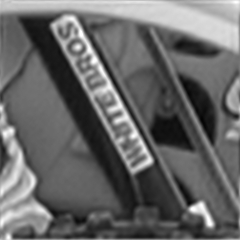}
	}\\
	\caption[Interpolation results of texture and text areas.]
			{{More interpolation results for subjective evaluation.}
			Results of 4x interpolation of LR inputs (a) texture and (b) text areas. 
			\vspace{-4mm}
			}
	\label{fig:upsres2}
\end{figure*}

Fig.~\ref{fig:upsres1} shows two input LR images, (a) and (b), and output HR images 
produced using directional, cubic spline and the proposed interpolation technique.
Directional interpolation results, (c) and (f), 
have some jaggedness for regions with strong edges and show some artifacts.
The cubic spline results, (d) and (g), do not have any strong artifacts but 
show blurring of edges. 
HR images produced using the proposed approach,
(e) and (h), 
are sharper and do not exhibit any noticeable artifacts.
Fig.~\ref{fig:upsres2} shows two more input LR images, (a) texture and (b) text areas, 
and their corresponding output HR images.
The texture in (e) appears slightly sharper than other methods,
and the text in (h) seems to be sharper than the other results.
It can also be seen that, even for intricate textures, the proposed method
produces results without evident artifacts.

%A subset of input and output images can be accessed at
%\small \url{http://iphome.hhi.de/lakshman/icip2013.html} \normalsize
%for subjective evaluation. 

One of the main drawbacks of the proposed approach is the high computational complexity.
The complexity of the proposed approach is much higher than that of typical FIR interpolators, 
but of the same order of magnitude as other non-linear methods such as 
the contourlet scheme~\cite{Mueller2007} and about 1.5x faster than the directional interpolation approach of~\cite{Li2001}.
Some important parameters that can be tuned for reducing the complexity are:
the number of iterations for sparse approximation, 
the number of scales, the number of orientations for the directional filtering, etc.
The filtering operations and element-wise thresholding involved in the proposed approach
are amenable to parallel implementation.

\section{Summary and discussion}
\label{sec:con}
\balance
The problem of image interpolation is closely related to image modeling, i.e.,
we ``select'' a particular HR image that fits our model 
from a set of images that satisfy the given LR data.
Unlike many other forms of data, images can show abrupt variations, e.g., across edges,
which introduces challenges in modeling.
In this paper, a framework for image interpolation that combines low frequencies from a linear method
and high frequencies from a sparse approximation was presented.
The key idea is to keep the support of the FIR filter short to avoid ringing artifacts in the initial upsampling
and attack the problem of blurriness of the resulting image using a high pass estimate, 
through a sparse approximation in a multi-resolution directional dictionary.

In this paper, we evaluated linear methods such as bicubic, 6-tap, 8-tap, and 12-tap filters, 
as well as spline based methods.
In the non-linear category, a directional interpolation method was evaluated, 
along with the proposed method using contourlet and shearlet dictionaries.
All the tested approaches perform well for smooth image regions, with the main differences
being observed at edges and in textured areas.
The linear methods have only a small number of free parameters and once a set of parameters
has been chosen, the performance variation from image-to-image is relatively small.
The non-linear methods have a higher number of free parameters,
hence a more careful setting is required.
Some quantitative methods were provided for parameter selection in the proposed approach.
With the final set of selected parameters, 
an average PSNR gain of around 0.63~dB was observed compared to a 8-tap
filter over a test set of 200 images. The maximum gain was around 3.13~dB, which is significant.
Additionally, many LR image regions with different characteristics were interpolated and subjectively evaluated.
The proposed method showed improvements in subjective quality compared to
other approaches and no evident artifacts were observed, 
even for complex regions.

\end{document}